\documentclass[sigconf,nonacm]{acmart}

\usepackage{array}
\usepackage{float}
\usepackage{enumitem}
\usepackage{makecell}
\usepackage{multirow}
\newlist{cellist}{itemize}{1}
\setlist[cellist]{
  leftmargin=0.85em,
  labelsep=0.3em,
  label=\textbullet,
  itemsep=0pt,
  parsep=0pt,
  topsep=0pt,
  partopsep=0pt,
  before=\vspace{-0.55\baselineskip},
}
\newcolumntype{L}[1]{>{\raggedright\arraybackslash}p{#1}}

\graphicspath{{figs/}}

\newcommand{\pipeline}{GenPHRI}

\definecolor{coral}{HTML}{c5544a}
\definecolor{blue}{HTML}{5392b4}
\definecolor{teal}{HTML}{479a9b}

\AtEndPreamble{%
  \hypersetup{
    colorlinks=true,
    linkcolor=coral,
    citecolor=teal,
    urlcolor=blue,
  }%
}

\begin{document}

\title[\pipeline: Agentic Generative Simulation for pHRI]{\pipeline: Agentic Generative Simulation\\%
for Physical Human-Robot Interaction}

\author{Junxiang Wang}
\authornote{Correspondence to \href{mailto:junxiang@cmu.edu}{\texttt{junxiang@cmu.edu}}}
\authornote{These authors contributed equally.}
\affiliation{
  \institution{Carnegie Mellon University}
  \city{Pittsburgh}
  \state{PA}
  \country{USA}
}

\author{Nir Pechuk}
\authornotemark[2]
\affiliation{
  \institution{Carnegie Mellon University}
  \city{Pittsburgh}
  \state{PA}
  \country{USA}
}

\author{Xinwen Xu}
\authornotemark[2]
\affiliation{
  \institution{Carnegie Mellon University}
  \city{Pittsburgh}
  \state{PA}
  \country{USA}
}

\author{Tiancheng Wu}
\authornote{These authors contributed equally.}
\affiliation{
  \institution{Carnegie Mellon University}
  \city{Pittsburgh}
  \state{PA}
  \country{USA}
}

\author{Nayon Kim}
\authornotemark[3]
\affiliation{
  \institution{Carnegie Mellon University}
  \city{Pittsburgh}
  \state{PA}
  \country{USA}
}

\author{Julian Millan}
\affiliation{
  \institution{Carnegie Mellon University}
  \city{Pittsburgh}
  \state{PA}
  \country{USA}
}

\author{Zackory Erickson}
\affiliation{
  \institution{Carnegie Mellon University}
  \city{Pittsburgh}
  \state{PA}
  \country{USA}
}

\renewcommand{\shortauthors}{Junxiang Wang et al.}

\begin{abstract}
  Real-world physical human-robot interaction (pHRI) development is costly and slow, while simulation requires substantial task-specific scene and motion authoring. We present \pipeline{}, a generative simulation framework for producing deployment-ready pHRI scenarios entirely from natural-language task descriptions using vision-language model (VLM) agents. Generated scenarios contain a room with furniture, a task-appropriate human pose and placement, and a robot executing the requested interaction. Generator and critic agents iterate on each component, and an orchestrator agent controls cross-stage decisions. With \pipeline, we generate 50 different assistive scenarios in simulation with no manual intervention and deploy two tasks in a 12-participant user study. Vision policies trained on these scenarios achieve zero-shot task completion rates of \textasciitilde80\% in real-world trials against \textasciitilde90\% in simulation, while producing behaviors that participants rate as consistent with textual task prompts. Further, we demonstrate direct execution of generated motion as a complementary, training-free deployment pathway that enables rapid real-world inspection of generated trajectories at the cost of contact coverage.
  We release \pipeline{} as a platform for generating, training, and deploying pHRI tasks from textual prompts. Project website: \url{https://rchi-lab.github.io/gen_phri/}.
\end{abstract}


\begin{teaserfigure}
  \includegraphics[width=\textwidth]{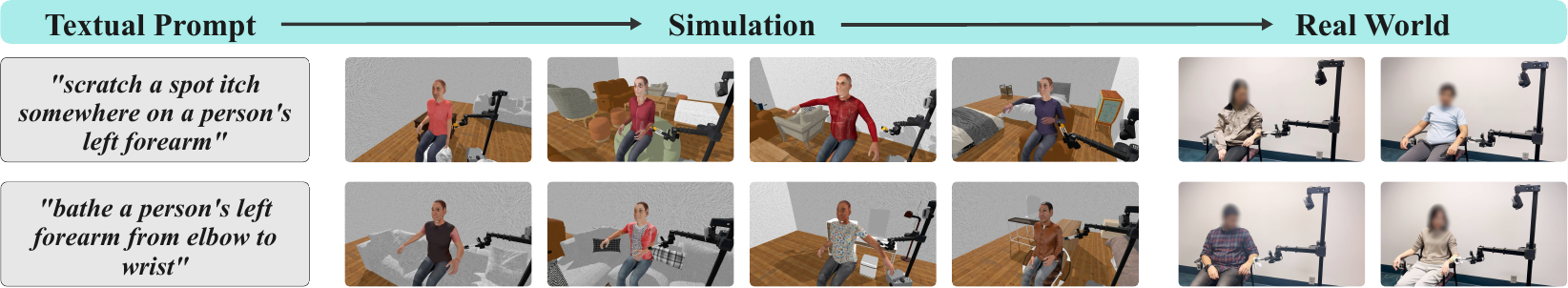}
  \caption{Overview of \pipeline{}, our proposed agentic generative simulation framework for synthesizing physical human-robot interaction (pHRI) scenarios. From a natural-language task prompt, \pipeline{} generates complete scenarios in simulation, which can be transferred zero-shot to real-world robot behaviors by directly executing the motion or training a policy.}
  \label{fig:teaser}
\end{teaserfigure}

\maketitle

\section{Introduction}

Robots that can autonomously make safe, task-productive physical contact with humans have the potential to unlock breakthroughs in healthcare, home assistance, collaborative manufacturing, and more. However, designing and iteratively improving these contact-rich behaviors in the real world is inherently slow and risky. While simulation offers a natural platform for rapid and safe iteration, existing physical human-robot interaction (pHRI) simulators are hand-designed for specific tasks~\citep{erickson2020assistivegym,ye2022rcare}, lacking task variability and scene diversity. Conversely, while recent advances in generative scene synthesis~\cite{wang2024architect,zhou2026virtual,ren2025simworld} have greatly improved the automated creation of diverse, photorealistic environments from text inputs, they do not address the unique challenges of pHRI, such as generating human body models and planning contact-rich robot motions to interact with human bodies.

To address this gap, we introduce \pipeline, an agentic generative simulation framework for synthesizing pHRI scenarios and approaches to transfer these scenarios into real-world robot behaviors (see Figure~\ref{fig:teaser}). From a single high-level task prompt, such as ``scratch a person's right forearm,'' \pipeline{} generates a complete environment, driven by vision-language model (VLM) agents. At each stage, generator agents propose human model parameters, spatial placements, robot configurations, and physically interactive motion trajectories, while critic agents verify each candidate in a closed loop. An overarching orchestrator adjudicates the metrics and findings from every stage to autonomously decide whether to proceed, retry, or backtrack. 

\pipeline{} provides two complementary methods for deploying generated scenarios onto real robots. For robust closed-loop control, \pipeline{} extends to collecting large-scale demonstration data across synthesized human variations and training zero-shot real-world-deployable visuomotor policies via imitation learning. Alternatively, for rapid deployment without data collection or model training, a body-mesh-fitting approach allows directly executing the generated motion on a real person, providing a fast, prompt-adhering physical impression of the task.

We demonstrate the scalability of \pipeline{} by generating 50 distinct assistive tasks in simulation without intervention. To evaluate real-world transfer, we deployed two tasks---back-and-leg bathing and forearm scratching---on a real mobile manipulator in a 12-participant within-subjects study. Our learned policies transferred zero-shot to the real world, achieving \textasciitilde80\% task completion. Participants evaluated the generated motions as adhering strongly to the initial textual prompt, demonstrating that \pipeline{} presents an opportunity to produce realistic and deployable physical interaction behaviors.

In summary, our primary contributions in this paper are:
\begin{itemize}[
  topsep=0pt,      
  partopsep=0pt,   
  parsep=0pt,      
  itemsep=0.1em,   
  leftmargin=1.5em,  
  labelsep=0.5em   
]
  \item \pipeline, the first agentic generative simulation framework for physical human-robot interaction (pHRI), using a generator-critic-orchestrator architecture to autonomously synthesize contact-rich tasks from natural-language prompts.
  \item Two approaches for transferring generated scenarios to real-world robot behaviors: policy learning over generated human variations, and training-free direct execution.
  \item A 12-participant user study on two assistive tasks with a real mobile manipulator, demonstrating \textasciitilde80\% zero-shot task success and strong adherence between robot behavior and task prompt.
\end{itemize}

\section{Related Work}

\subsection{Physical Human-Robot Interaction}
A primary application of physical human-robot interaction (pHRI) is assisting people with activities of daily living, including feeding~\cite{jenamani2024feelthebite}, bathing~\cite{madan2024rabbit,gu2024vttb,gu2024learning}, and dressing~\cite{clegg2020collaborate,hao2025forcemodulated,sun2026dressingmotion}. These tasks require purposeful contact with the body while accommodating differences in anatomy, mobility, and behavior. Consequently, developing an assistive system involves both perceiving the person and coordinating robot motion with the geometry and physical constraints of the interaction.

Simulation provides a controlled environment for investigating these challenges before deploying robots around people. AssistiveGym and RCareWorld offer human models, contact dynamics, and caregiving benchmarks~\cite{erickson2020assistivegym,ye2022rcare}, while Human-Robot Gym supports safety-constrained collaboration~\cite{thumm2024humanrobotgym}. Assistive VR Gym further incorporates real user responses to improve simulated human models~\cite{erickson2020assistivevr}. These platforms support variation in human characteristics and movement, and simulation-trained dressing controllers demonstrate transfer despite deformable garments and moving arms~\cite{clegg2020collaborate,hao2025forcemodulated}. However, constructing the task setup and specifying the intended interaction remain substantial authoring requirements. Extending a simulator to a new caregiving request therefore requires selecting a suitable posture, arranging the scene, and defining the robot's contact behavior.

Recent work addresses personalization through functional-range-of-motion models~\cite{liu2025grace}, biomechanically constrained limb repositioning~\cite{koh2025manip4care}, and context-aware adaptation across environments and robot embodiments~\cite{wu2026embodiment}. Expert datasets and contact-preserving demonstration transfer also provide rich examples of caregiving behavior~\cite{liang2025openrobocare,lakshmipathy2026bathing}. These advances improve how robots perform assistance once relevant skills or demonstrations are available. \pipeline{} addresses the complementary problem of creating the interaction itself: generating task-appropriate humans, compatible placements, and contact trajectories from text, with no task demonstrations required for either direct motion execution or policy training.

\subsection{Generative Simulation}
Generative simulation offers a route to reducing the manual effort required to construct diverse interaction environments. Text- and image-conditioned methods synthesize room geometry and spatial layouts~\cite{hoellein2023text2room,yang2024holodeck,wang2024architect}, with recent platforms extending generation to broader physical and social worlds~\cite{ren2025simworld,zhou2026virtual}. Human-motion generation additionally addresses scene grounding and physical plausibility~\cite{wang2022humanise,xiao2024unihsi,yuan2023physdiff}. Although these methods provide useful ingredients for human-centered environments, generating plausible surroundings and human behavior does not establish how a robot should make task-appropriate contact with a person.

Other approaches connect generation more directly to robot behavior by synthesizing tasks and simulation programs~\cite{wang2023gensim,wang2023robogen,hua2025gensim2}, designing rewards and randomization~\cite{ma2023eureka,ma2024dreureka}, or expanding seed demonstrations~\cite{mandlekar2023mimicgen,xue2025demogen,nasiriany2026robocasa365}. Behavior-conditioned scenario generation and video-world-model trajectories offer further avenues for scaling experience~\cite{nguyen2025regen,jang2025dreamgen}. Together, these methods automate substantial parts of environment and behavior creation, motivating their extension to the coupled requirements of physical assistance.

\begin{figure*}[t]
  \centering
  \includegraphics[width=1.0\linewidth]{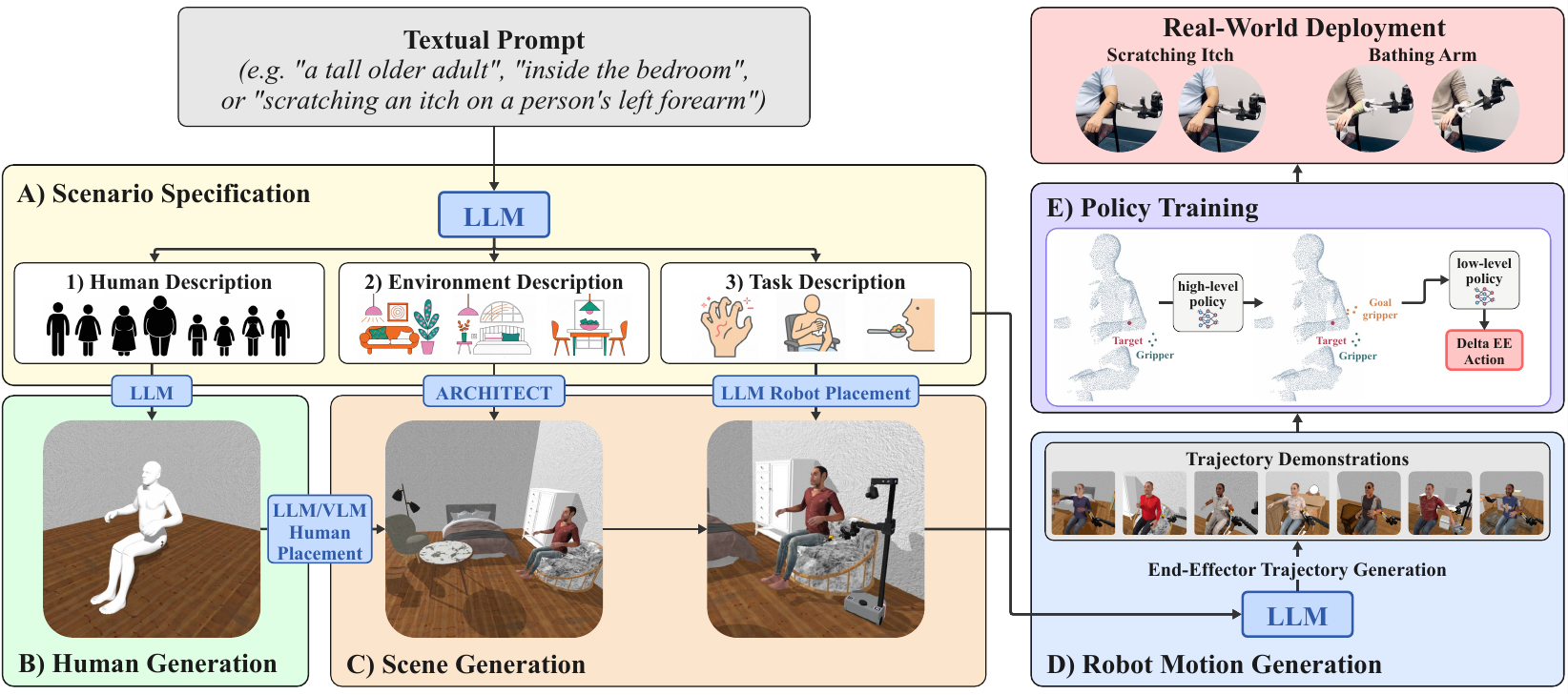}
  \caption{Flowchart of \pipeline, an agentic generative simulation system for physical human-robot interaction. An orchestrator agent takes in a task prompt and sends detailed prompts to four generation stages that compose a scenario, including a posed human and robot within a scene and task-completing robot motion. The resulting scenario can be either transferred directly into the real world through mesh fitting and motion planning, or used in training a policy with zero-shot sim-to-real capabilities.} 
  \label{fig:main}
\end{figure*}

Closely related, SceneSmith uses designer, critic, and orchestrator agents to iteratively construct simulation-ready indoor scenes~\cite{pfaff2026scenesmith}. We extend agentic generative simulation to pHRI, where human pose, support, robot placement, and contact behavior must be mutually compatible. A plausible human placement, for example, may prevent the robot from reaching the requested body region. \pipeline's cross-stage feedback revisits such decisions using visual critique and simulator checks, producing complete assistive scenarios for direct execution or policy learning.

\section{Agentic pHRI Scene Construction}
\label{sec:methods}

\label{sec:system-overview}
We introduce \pipeline, an agentic generative simulation framework for synthesizing physical HRI scenarios, as shown in Figure~\ref{fig:main}. Given an initial task prompt, each scenario is generated in four stages: human generation (\S~\ref{sec:human-gen}), human placement (\S~\ref{sec:human-placement}), robot placement (\S~\ref{sec:robot-placement}), and robot motion generation (\S~\ref{sec:motion-gen}). An orchestrator (\S~\ref{sec:orchestrator}) oversees the generation process, operating over the four stages. The orchestrator generates new indoor environments using SceneSmith~\citep{pfaff2026scenesmith} or selects from their scene bank, and we simulate the resulting pHRI scenarios in Genesis~\citep{Genesis}. Generated scenarios support direct deployment through real-world human-mesh fitting (\S~\ref{sec:sam3d}) or policy learning (\S~\ref{sec:policy}). For policy learning, orchestrator oversight continues through data collection, training, and simulated evaluation.

\subsubsection{Problem Formulation}
Given a fixed task specification $\tau$, we seek a scenario $x = (A_{\mathrm{room}}, A_{\mathrm{pose}}, A_{\mathrm{position}}, A_{\mathrm{robot}}, A_{\mathrm{motion}})$. Assignments $A_{\mathrm{component}}$ represent room configuration, human pose, human-placement parameters, robot-placement code, and motion-generation code respectively. $\tau$ is generated once at the start of the run and includes information on the human posture, motion type, and supporting furniture.

Based on this scenario representation, we formulate generation as a utility maximization problem under constraints:
$$
\underset{x \in \mathcal{X}_{\tau}}{\operatorname{maximize}}
\; U_{\tau}(x)
\quad \text{subject to} \quad
c_j(x ; \tau)=1,\quad j=1,\ldots,m,
$$
where $c_j$ is a required acceptance condition over components of $x$, and $U_{\tau}$ represents task-quality preferences. For example, waypoint reachability is a constraint dependent on human pose, placement, robot placement, and motion. We focus on manipulation over navigation, making the assumption that robot placement stays fixed during motion.

\subsubsection{Scenario Optimization}
We utilize a two-layer design to separate local candidate refinement from search over upstream alternatives. First, the orchestrator (\S~\ref{sec:orchestrator}) converts the user's short-form task description into the structured task specification $\tau$, which it uses to supply task prompts to generators and critics. Through a generate-render-critique loop (\S~\ref{sec:scenario-gen}), each stage produces an assignment to one component of the scenario with previous assignments fixed. We separate generation from critique so that acceptance is assessed against the original task description by a distinct evaluator, aiming to limit task drift and generator self-acceptance. Between stages, the orchestrator uses cross-stage feedback to select which assignments to revisit (\S~\ref{sec:orchestrator}).

At the beginning of each loop, the generator is supplied context on its role by the orchestrator and interacts with a simulator API to create an assignment $A_\mathrm{component}$ for the scenario. The simulator renders it and computes stage-specific metrics: gates enforce required conditions $c_j$, while advisory flags inform quality judgments associated with $U_{\tau}$. For example, motion waypoints must be reachable, whereas human self-collision contact pairs may be acceptable if penetration is small and visually plausible. Next, the critic evaluates renderings and metrics against the fixed task specification $\tau$ and stage-specific criteria, including judgments beyond the metrics, such as whether the body-furniture relation matches the task. Stage-specific renderings and criteria appear in \S~\ref{sec:scenario-gen} and Appendix~\ref{sec:stage-breakdown}. The critic returns an overall severity (negligible, mild, moderate, or severe) and findings containing issue summaries, suggested changes, and rough magnitudes. These findings condition the generator's next candidate. The loop exits when the critic accepts and every gate passes, or when its iteration budget is exhausted. The project website contains example revisions for each stage.

After each loop, the orchestrator receives the final candidate's metrics and critique. If the final candidate was a failure, it also receives the best-rated candidate. It chooses to accept the final or best candidate, retry the current stage, or backtrack. Backtracking reopens upstream assignments when local refinement cannot resolve cross-component violations: for example, failed contact motion may require moving the robot closer to the human. 

\subsection{Generator-Critic Component Creation}
\label{sec:scenario-gen}

\subsubsection{Human Generation}
\label{sec:human-gen}

This stage generates a simulated human with a task-appropriate posture and diverse body shape. We represent the human using the SMPL-X~\cite{SMPL-X:2019} parameterized body model, whose function $M(\theta, \beta)$ produces a 3D body mesh from posture parameters $\theta$ and shape parameters $\beta$. This stage takes a high-level description of the desired posture and body shape from the orchestrator and outputs the corresponding SMPL-X parameters through a two-part generation scheme. First, the generator queries an LLM to expand the high-level human description into detailed semantic descriptions of body shape and body pose (done separately). Next, these detailed descriptions are mapped into concrete numerical values through in-context learning by providing example pairings of pose descriptions with human joint values. This two-part design grounds the generated parameters in the semantic context of the SMPL-X model to promote consistency with the input description. For actuation, we construct a human model URDF with its zero configuration as the generated posture, and for improved visual realism, the generated human mesh is textured using the SMPLitex dataset~\cite{casas2023smplitex}.

Then, the critic visually verifies the generated human body mesh through image renderings from two different angles and checks for implausible joint configurations and self-intersections. If issues are found, the critic provides feedback for the generator to improve the assignment.

\subsubsection{Human Placement}
\label{sec:human-placement}
This stage places the generated human given orchestrator-supplied furniture, indoor environment, and desired posture. From a top-down rendering of the furniture's axis-aligned bounding box with specified coordinate conventions (see Figure~\ref{fig:main}), the generator produces two normalized coordinates $x,y\in[0,1]$ and two directions $d_{\text{head}},d_{\text{face}}\in\{\text{up}, \text{down}, \text{left}, \text{right}, \text{in}, \text{out}\}$ to define the human's horizontal position and 3D orientation. Here, $d_{\text{head}}$ denotes the direction from torso center to head, and $d_{\text{face}}$ denotes the direction from back to chest. For seated poses, the human's vertical position is computed as the contact height at which the human and supporting furniture meshes touch, given the generated posture, orientation, and horizontal position. If they are in a lying pose, we allow the actuated human URDF to settle on the furniture and update the body parameters to ensure a natural pose.

The critic inspects image renderings of the settled human on the furniture from a top-down and isometric view. It specifically checks for any intersection between the human body and the furniture, and whether the relation between the two contradicts the input prompt. Only seated and lying postures are currently supported, as these are most common in assistive scenarios.

\subsubsection{Robot Placement}
\label{sec:robot-placement}
This stage positions and orients the robot relative to the placed human to support the requested interaction. The generator accesses the human's pose through a simulator API that returns the 3D positions of skeletal joints, including wrists, elbows, shoulders, neck, hips, knees, and ankles. Given an orchestrator prompt describing where the robot should be positioned next to the human, it produces Python code defining the robot's position and orientation. Restricting the generated code to this single simulator API enforces that the robot's pose is computed from human joint positions instead of in absolute coordinates.

\begin{figure}[tb]
  \centering
  \includegraphics[width=\linewidth]{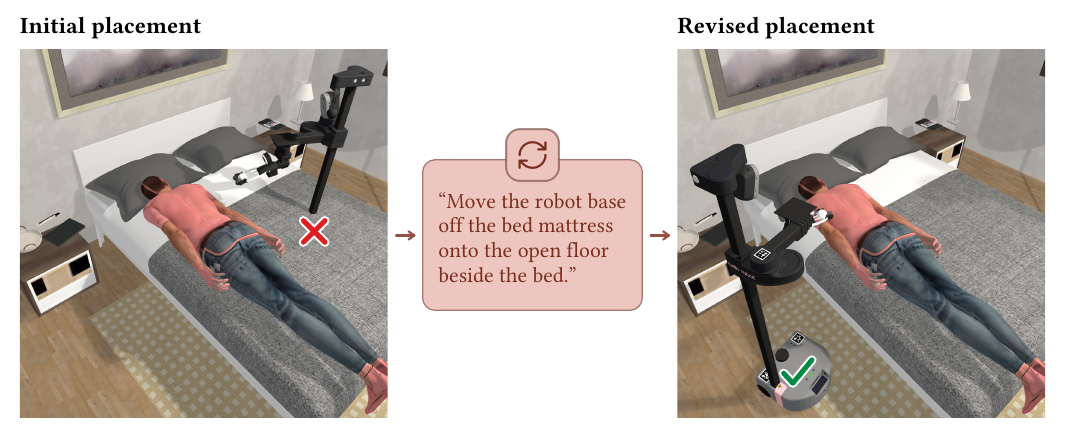}
  \caption{Robot placement before (left) and after (right) a critic iteration. The first candidate places the robot intersecting the furniture, while the revised placement is valid.}
  \label{fig:critic}
\end{figure}

The critic receives this code alongside top-down and isometric renderings of the human and robot within the scene. It evaluates whether the placement supports the requested task, checking collisions with surrounding objects, proximity to the human, robot orientation, and overall task feasibility. See Figure~\ref{fig:critic} for an example where the critic corrects an initial generator candidate robot placement.

\subsubsection{Robot Motion Generation}
\label{sec:motion-gen}

\begin{figure}[tb]
  \centering
  \includegraphics[width=\linewidth]{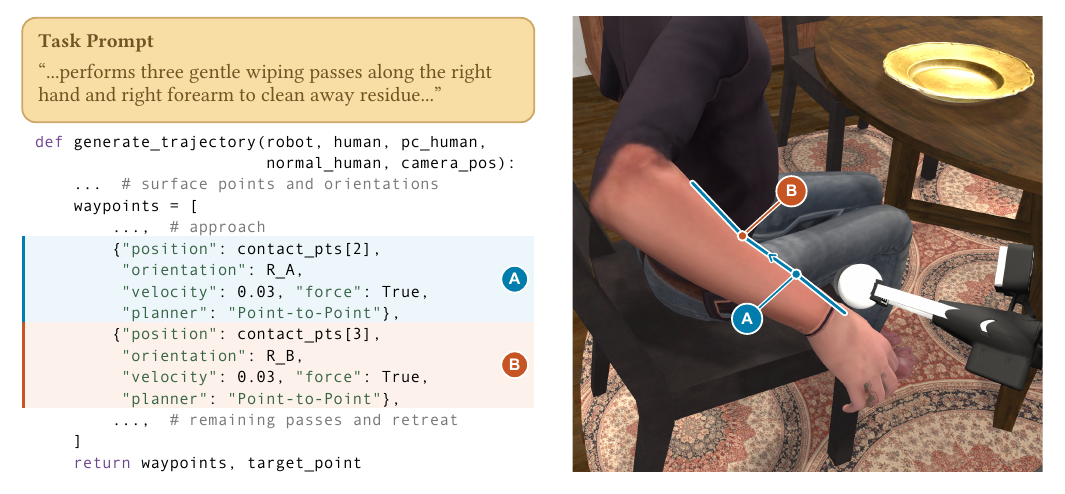}
  \caption{Robot motion generation. Left: excerpt from task prompt and generated trajectory code. Right: visualization of the resulting motion trajectory.}
  \label{fig:motion-gen}
\end{figure}

This stage generates a robot motion trajectory for the requested physical interaction. Given the task description, the generator produces code defining a function that computes a trajectory from human skeletal joint positions and other quantities (see Figure~\ref{fig:motion-gen}). The inputs to the function include a human point cloud, per-point surface normals, and the camera pose. Within the function, APIs are available for querying human skeletal joint positions and projecting these joints or any 3D point onto the observed point cloud surface for contact accuracy. These input quantities and APIs are designed to allow the generator to map skeletal landmarks, such as the wrist or elbow, or interpolations between these landmarks, to corresponding locations on the visible body surface, with orientation alignment from surface normals.

The generated function outputs a sequence of waypoints; each specifies the desired end-effector position and orientation, velocity, planned contact, and the planning algorithm used to reach it from the preceding waypoint. The available algorithms are a rapidly-exploring random tree (RRT) planner for collision-free motion through free space and a straight-line Cartesian planner for close-proximity and contact-rich phases. The input quantities and APIs ground the generated waypoints in human body parts and joints rather than fixed world coordinates.

Unlike the preceding critics, the motion critic receives a video of the simulated robot executing the trajectory, alongside renderings overlaid with the planned waypoints. This supplies temporal evidence of the executed interaction alongside spatial views of the intended trajectory. From the video and images, the critic assesses consistency with the task description and whether the tool contacts the intended body regions. It also inspects the code and numerical metrics for appropriate contact flags, prescribed contact velocities, end-effector orientation, and waypoint reachability. These complementary inputs support assessment of both the requested behavior and the feasibility of its execution.

\subsection{Orchestrator Agent}
\label{sec:orchestrator}
Generator-critic loops refine each component of the pipeline in isolation. This can leave a stage unable to resolve a problem caused by an earlier choice. For example, if the robot is placed too far from the target body region, refining the motion alone may not make the interaction reachable. We introduce an orchestrator that can revisit earlier choices to resolve such incompatibilities and improve scenario quality. The project website contains detailed examples of orchestrator outputs.

\subsubsection{Task Expansion}
At the start of the run, the orchestrator expands the user's high-level pHRI task description into $\tau$, specifying the task, human posture, supporting furniture types, target body segments, and motion pattern. This specification remains fixed throughout generation.

Before each stage, the orchestrator prepares generator instructions from a template for the stage, editing it based on previous stage outputs and their critics' feedback. This generator's critic still evaluates the resulting artifacts against the original task and $\tau$ without orchestrator changes, so generators can adapt independently of task requirements. Otherwise, the critic could weaken a requirement to accommodate a shortcoming rather than proposing changes that satisfy the original request.

\subsubsection{Cross-Stage Coordination}
After stage completion, failure, or budget exhaustion, the orchestrator selects among three actions: \texttt{accept}, \texttt{steer}, or \texttt{backtrack}. The chosen action is checked against deterministic conditions such as remaining budget and allowed transitions.

The orchestrator can \texttt{accept} an eligible best or latest candidate and proceed to the next stage. When local changes may resolve the problem or improve quality, it can \texttt{steer} by resuming or retrying the current generator-critic loop, optionally providing revised instructions. When feedback points to an upstream choice that limits the current stage, it can \texttt{backtrack}, revisiting that component with revised generator guidance while preserving $\tau$. Generation ends when the pipeline completes.

\subsubsection{Artifact Invalidation}
When backtracking, we use an artifact dependency graph to regenerate affected downstream results while preserving reusable results. For example, changing the human pose can require updating placement and motion, whereas changing only the robot motion preserves the existing human and robot placements.

\subsubsection{Context Management}
At each invocation, the orchestrator receives a condensed summary of the current pipeline state with relevant assignments, stage outcomes, deterministic metrics, remaining budgets, and recent decisions. We prioritize including failed constraints that could inform actions. Restraining the context window allows the orchestrator to focus on the most relevant details, improving performance.

\subsection{Transfer to the Real World}
\label{sec:real-world-transfer}

\subsubsection{Direct Execution}
\label{sec:sam3d}

Direct execution provides a lightweight route to real-world robot behaviors without data collection or policy training, enabling rapid iteration on generated motion. The human, robot, and camera are arranged similarly to the simulated scenario, with the camera calibrated relative to the robot. From a single RGB image, Fast SAM3D Body~\citep{yang2026fast} estimates an SMPL-X mesh, including body shape $\beta$, posture $\theta$, translation, and orientation. The fitted skeletal joint positions, mesh vertices and surface normals, together with the calibrated camera pose, supply the inputs required by the generated robot motion function. The resulting trajectory is executed by the robot's low-level controller.

Execution quality, particularly during contact, depends on the accuracy of the fitted mesh. We use an open-loop procedure, only fitting a single mesh on the starting image to plan the trajectory. Closed-loop operation could improve performance but would require updating the body estimate despite occlusions introduced by the robot during interaction.

\subsubsection{Policy Learning}
\label{sec:policy}
For closed-loop execution, \pipeline{} supports large-scale demonstration collection, policy training, and simulated policy evaluation as optional pipeline stages coordinated by the orchestrator. By instantiating diverse human bodies and augmenting the generated demonstrations, we train policies that respond to visual observations throughout the interaction. These policies support zero-shot real-world deployment and provide opportunity for future work on subsequent task-specific fine-tuning.

We train vision-based policies with human point clouds as our primary visual representation, adapting the hierarchical imitation learning policy architecture introduced in Articubot~\cite{wang2025articubot}. In this architecture, a high-level policy with a PointNet++~\citep{qi2017pointnet++} backbone takes a point cloud of interest $P\in\mathbb{R}^{N\times 3}$ and four points representing the current robot end-effector gripper pose $C = \{c_i\}_{i=1}^4$ as input, and outputs a goal pose of the robot end-effector $G = \{g_i\}_{i=1}^4$ through a weighted displacement model. Subsequently, a low-level 3D Diffusion Policy~\cite{ze2024d} conditions on this predicted goal to generate step-by-step actions (delta end-effector poses) to reach that goal. In our context, the input point cloud is the point cloud of the human, and the high-level policy prediction target represents the next waypoint along the planned trajectory.

This architecture makes the assumption that the high-level goal $G$ can be uniquely identified from only the point cloud $P$ and the current end-effector pose $C$; e.g., for a drawer-opening task, if the end-effector is at the handle, then the goal pose would be the final position after the opening motion. However, this assumption does not hold for many pHRI scenarios; e.g., when massaging back and forth along an arm, knowing that the end-effector is in the middle of the arm does not uniquely identify the next waypoint, since the motion can be either forward or backward. To adapt the architecture in these multimodal scenarios, we change the high-level policy backbone from PointNet++ to Point Transformer V3~\citep{wu2024point}, which allows the prediction to be conditioned on a discrete variable $I$. For a generated trajectory of $K$ waypoints, each waypoint can then be associated with a unique condition integer in $[0,K-1]$ representing its order in the trajectory. At inference time, this condition integer is instantiated to $I=0$ and incremented whenever the current end-effector pose is within a distance threshold from the predicted goal pose:
\begin{equation*}
  \frac{1}{4}\sum_{i=1}^4 \|c_i-g_i\|_2 < d_{\text{thresh}}
\end{equation*}
This policy-learning approach predicts the next waypoint and delta actions towards the waypoint directly from point cloud observations without mesh fitting. The policy is robust to occlusions since the training data naturally include occlusions as robot motions are executed in simulation.

\paragraph{Policy-Training Pipeline Extension} Data collection, training, and simulated evaluation are integrated into the pipeline as individual stages for policy-based deployment. The orchestrator treats high- and low-level training as separate stages, checking completion and numerical stability and reviewing validation-loss trends and episode-level outliers. These reports assess training quality, while subsequent policy rollouts on training and held-out human bodies assess task performance. Depending on the findings, the orchestrator can select an earlier checkpoint, revise training settings, or backtrack to data collection or scenario generation. Appendix~\ref{sec:stage-breakdown}
(Table~\ref{tab:policy-stage-reports}) summarizes the corresponding
gates and flags.

\section{Scaling Up Generative Simulation for pHRI}
\label{sec:sim}

To demonstrate the variety and coverage of the robot motions that \pipeline{} produces, we generated 50 scenarios across 9 task families. Scenarios in the same family target different areas of the human body (e.g., massage of the arm versus the leg). We use Gemini 3.7 Flash for all LLM and VLM calls in \pipeline, but this is an easily configurable pipeline parameter. Figure~\ref{fig:teaser} shows snapshots of sample scenarios. Appendix~\ref{sec:sim-details} gives an overview of all 50 scenarios, and our project website shows each scenario interactively. Table~\ref{tab:critic-orchestrator} counts the critic iterations and orchestrator decisions in the process of generating each scenario.

\begin{table}[t]
\centering
\caption{Critic and orchestrator activity in generating the 50 scenarios. Stages are: 1. Human generation, 2. Human placement, 3. Robot placement, and 4. Robot motion generation.}
\label{tab:critic-orchestrator}
\begin{tabular}{cccc}
\toprule
Stage & \makecell{Scenarios with\\$\geq$1 critic rejection} & \makecell{Mean critic\\iterations} & \makecell{Orchestrator\\decisions} \\
\midrule
1  & 22/50 & 2.02 & 5  \\
2   & 40/50 & 3.22 & 7  \\
3   & 14/50 & 4.16 & 25 \\
4      & 37/50 & 6.62 & 18 \\
\midrule
Total             & 50/50 & 16.0   & 55 \\
\bottomrule
\end{tabular}
\end{table}

\paragraph{Critics}
All generated scenarios required closed-loop critic feedback in at least one of the generation stages, averaging 16 total iterations across all stages. Rejections most frequently occur in human placement (40/50) and robot motion generation (37/50). Human placement, which uses a VLM generator, reflects the inherent visual grounding challenges in rendered scenes. Robot motion generation took the most iterations on average (6.62), as it requires enforcing challenging physical constraints like end-effector orientation alignment, contact quality, and motion reachability.

\paragraph{Orchestrator}
When a critic loop uses its full iteration budget without accepting a result, the orchestrator selects a recovery action. Overall, 16 of the 50 generated scenarios required orchestrator intervention, resulting in 55 total recovery decisions: 42 retries with injected context and 5 local budget extensions (\texttt{steer}), 3 \texttt{backtrack}s to an upstream stage, and 5 \texttt{accept}s of the best iteration. Backtracking prevented cascading failures by revising upstream placements when motion generation could not resolve waypoint reachability constraints.

\paragraph{Isolating Contributions}
We evaluate the impact of the orchestrator and each critic agent in a 7-way ablation. We run the back-and-leg bathing task, chosen as a representation of long-context full-body motion, with seven configurations: the full pipeline, critics only (no orchestrator), four variants that additionally remove one stage's critic, and fully open-loop generation. Single-critic ablations isolate effects of local refinement, while comparing critics only with the full pipeline examines the benefit of revisiting upstream decisions. Each configuration contributes a scenario evaluated across its cohort of 100 human bodies.

Table~\ref{tab:motion-ablation} reports all-waypoint reachability, which jointly constrains the human configuration, robot placement, and all sections of the robot trajectory. A locally plausible component can restrict subsequent stages, making this motion gate a concrete measure of cross-component compatibility. Because this measure is seen by the orchestrator during generation, we complement it with execution measurements not supplied as numerical feedback: robot end-effector deviations above 2\,cm from the planned trajectory (tracking excursions) and 95th-percentile distance of the robot's end effector from the planned endpoint (P95 final position error). Tracking excursions capture deviations that may compromise safety near the human and final position error measures whether the configuration allows the robot to reach its intended goal.

Against critics only, the full framework increases all-waypoint reachability from 57\% to 100\%, reduces tracking-excursion incidence from 4\% to 0\%, and lowers P95 final-position error from 152.3\,mm to 1.8\,mm. Removing any individual critic further reduces reachability and motion safety, while open-loop generation fully fails to produce plausible scenes. 

These results show the value of agentic cross-stage orchestration and closed-loop component iteration for pHRI scenario generation.

\begin{table}[tb]
  \centering
  \caption{Reachability and target-tracking metrics for the pipeline ablation.}
  \begin{tabular*}{\columnwidth}{@{\extracolsep{\fill}}lccc@{}}
    \toprule
    Configuration
    & \makecell{All waypoints\\reachable (\%) $\uparrow$}
    & \makecell{Tracking\\excursions\\$>$2\,cm (\%) $\downarrow$}
    & \makecell{P95 final \\ pos. error \\ (mm) $\downarrow$} \\
    \midrule
    Full pipeline    & 100.0 &   0.0 &   1.8 \\
    Critics only     &  57.0 &   4.0 & 152.3 \\
    \midrule
    \multicolumn{4}{@{}l}{\textbf{Critics only, minus:}} \\
    Human Generation &   0.0 & 100.0 & 155.9 \\
    Human Placement  &  37.0 &  50.0 &  77.4 \\
    Robot Placement  &   1.0 &  81.8 &  69.9 \\
    Motion Generation&   0.0 & 100.0 & 330.0 \\
    \midrule
    Open-loop        &   0.0 & 100.0 & 349.3 \\
    \bottomrule
  \end{tabular*}
  \label{tab:motion-ablation}
\end{table}

\section{Zero-Shot Real-World Deployment}
\subsection{User Study}
\label{sec:study}

We evaluate \pipeline's two deployment pathways in a within-participants study ($N=12$, 6\,male/6\,female, age range 18--35 with $M=22$ and $SD=5.5$), approved by our Institutional Review Board. The evaluation addresses three questions:

\textbf{RQ1}: Do policies trained on generated demonstrations transfer to real-world pHRI tasks without fine-tuning?

\textbf{RQ2}: Do generated and learned behaviors preserve the intent of the task description?

\textbf{RQ3}: How do direct execution and policy learning support different deployment needs?

\subsubsection{Tasks and Conditions}
\begin{figure}[tb]
  \centering
  \includegraphics[width=\linewidth]{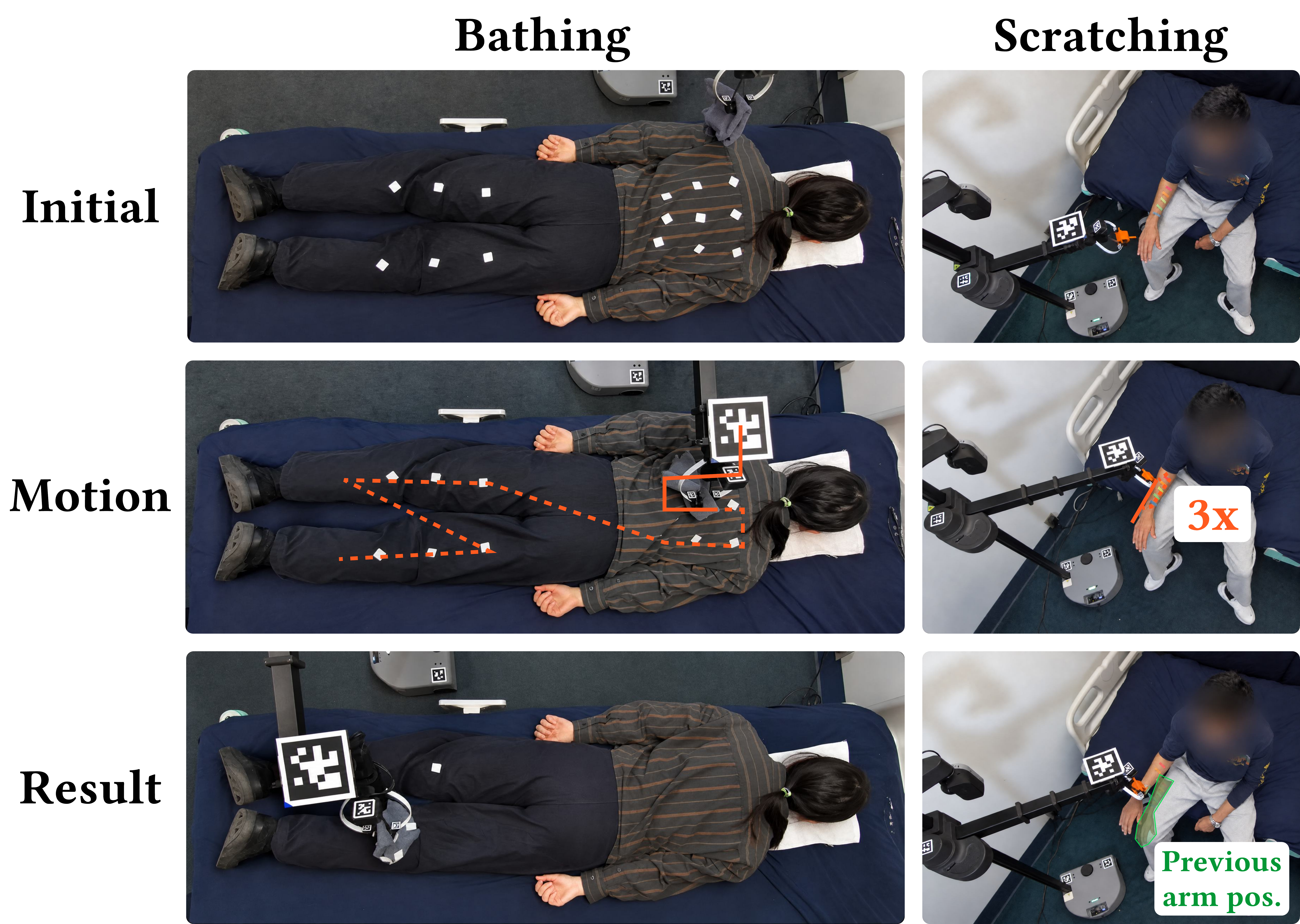}
  \caption{Real-world configurations for bathing and scratching, with visualizations of robot motion, final outcome, and human arm movement in scratching.}
  \label{fig:realworld}
\end{figure}

We deploy two generated assistive scenarios: \emph{bathing}, in which the robot wipes the participant's back and the backs of their legs with a towel, and \emph{scratching}, in which it makes three passes along the forearm from elbow to wrist (Figure~\ref{fig:realworld}). These tasks emphasize complementary interaction requirements: bathing involves extended motion across multiple body regions, whereas scratching requires localized contact along the forearm.

Bathing compares a learned policy with direct motion execution. Scratching compares a policy with and without a one-time arm displacement during the second pass, where the displacement magnitude is chosen by the participant for naturally varying movements. The arm-movement condition serves to test the policy's ability to adapt to changes in human pose when not explicitly trained on it. Each condition is repeated twice, yielding eight physical trials per participant. Each policy is trained on 1,000 simulated demonstrations across 100 human bodies and deployed without real-world fine-tuning. Hardware and training details appear in Appendices~\ref{sec:hardware} and~\ref{sec:policy-details}.

\subsubsection{Measures and Procedure}
We measure physical task performance through \emph{target completion}: the fraction of 15 evenly distributed velcro squares removed from the back and legs during bathing, or five adhesive markers contacted along the forearm during scratching. After each trial, participants view the original task prompt and rate how closely the experienced interaction follows it on a seven-point scale, with seven indicating the highest adherence. We refer to this measure as \emph{prompt adherence}. The prompts can be found in Appendix~\ref{sec:task-prompts}.

In a separate set of trials, participants answer the same prompt-adherence question for four simulation videos: two generated demonstrations and two learned-policy rollouts for each task. We counterbalance the order of the physical and video trial sets, the two tasks, and conditions within each set. Participants are not informed of the underlying deployment or generation methods.

\paragraph{Analysis}
For prompt-adherence comparisons, we aggregate repeated trials into a median per participant and condition, and use two-sided paired Wilcoxon signed-rank tests ($n=12$). To assess preservation of task intent, we test equivalence between generated demonstrations and simulated policy rollouts using a $\pm1$-point margin, and non-inferiority of scratching adherence under arm movement using a one-point margin. We compare bathing target completion between deployment methods using a paired $t$-test. Reported $p$-values are unadjusted.

\subsubsection{Simulated Evaluation}
We evaluate each policy on 20 held-out human bodies with two random seeds per body, yielding 40 trials per condition. Target arrangements mirror the physical study: 15 targets across the back and legs for bathing and five along the right forearm for scratching. In the movement condition, we perturb arm joints during the second pass to produce approximately 2--3\,cm of displacement. A simulated target is completed when the tool passes within 2\,cm of its position.

\subsection{Results}
\label{sec:real-results}

\subsubsection{Zero-Shot Physical Task Performance}
Policies trained exclusively on generated demonstrations achieve mean real-world target completion of 80\% for bathing and 79\% for scratching without arm movement (Table~\ref{tab:success}). Corresponding simulated completion is 86\% and 92\%. These results answer \textbf{RQ1}: policies trained only on generated demonstrations transferred to both real-world tasks without fine-tuning, showing that GenPHRI can provide simulation-generated training data for physical interaction with real humans.

With instructed arm movement, scratching completion decreases to 63\% in real-world trials, compared with 92\% under simulated perturbations. The policy therefore retains some ability to complete the task when the participant moves, but the gap between simulated and real-world performance indicates room for improved generalization, for example through more realistic simulated human motion or real-world fine-tuning.

\begin{table}[tb]
  \centering
  \caption{Mean target completion (\% $\pm$ SD across humans) in physical and simulated evaluations. Physical targets require removal or contact. Simulated targets use a 2\,cm proximity criterion.}
  \begin{tabular}{c|cc|cc}
    \toprule
    & \multicolumn{2}{c|}{\textbf{Bathing}}
    & \multicolumn{2}{c}{\textbf{Scratching}} \\
    & Policy & Direct & Policy & Policy+Motion \\
    \midrule
    Real & $80 \pm 11$ & $39 \pm 25$ & $79 \pm 16$ & $63 \pm 20$ \\
    Sim  & $86 \pm 7$  & ---         & $92 \pm 14$ & $92 \pm 20$ \\
    \bottomrule
  \end{tabular}
  \label{tab:success}
\end{table}

\subsubsection{Preserving Task Intent}
\begin{figure}[tb]
  \centering
  \includegraphics[width=\linewidth]{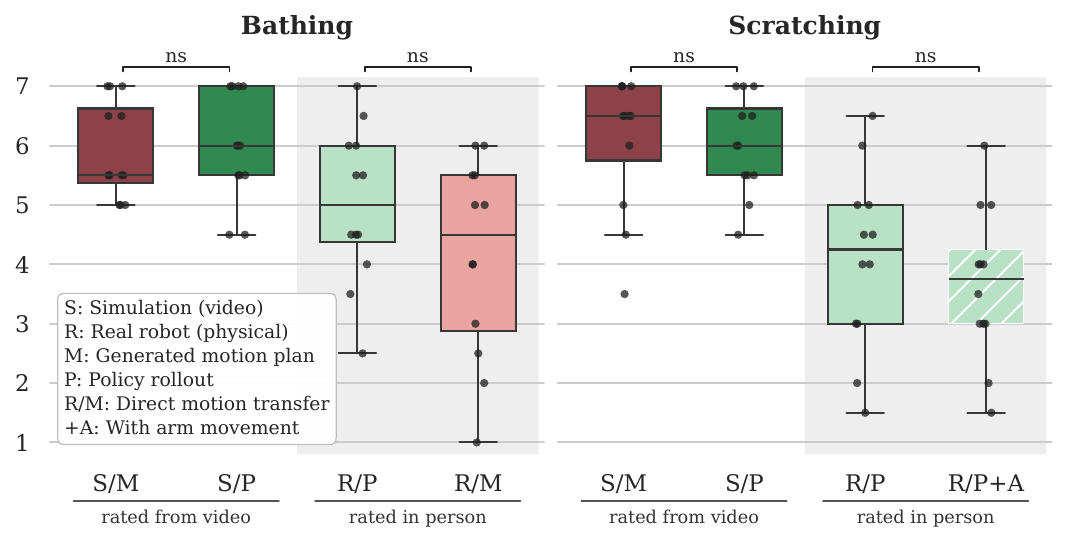}
  \caption{Participant ratings of prompt adherence. ``ns'' denotes a nonsignificant difference in the two-sided paired comparison.}
  \label{fig:adherence}
\end{figure}

To address \textbf{RQ2}, we first examine prompt adherence in simulation, where participants can directly compare generated demonstrations with learned policy rollouts. Both are rated highly, with median ratings between 5.5 and 6.5 on the seven-point scale (Figure~\ref{fig:adherence}). Ratings for generated demonstrations and policy rollouts are equivalent within our one-point margin for both tasks ($p<0.01$), indicating that policy learning does not meaningfully reduce perceived adherence to the original task prompt.

We then examine whether this adherence carries over to physical interaction. Real-world policy executions receive median ratings of 5.0 for bathing and 4.25 for scratching, both above the midpoint of the scale. For scratching, adherence under instructed arm movement is non-inferior to adherence without movement within our one-point margin ($p<0.01$). Together, these results answer \textbf{RQ2}: generated behaviors are perceived as consistent with their task prompts, and this task intent is largely preserved through policy learning and during real-world interaction, including when the participant moves.

\subsubsection{Comparing Deployment Pathways}
\label{sec:sam3d-discussion}
For bathing, direct execution receives a median prompt-adherence rating of 4.5, compared with 5.0 for the learned policy; this difference is not statistically significant. However, target completion is substantially lower with direct execution, at 39\%, compared with 80\% for policy learning ($p<0.001$). 

The lower contact coverage of direct execution is primarily caused by errors in estimating participant depth from a single RGB image. These errors can shift the fitted body mesh by several centimeters, causing the robot to execute the intended motion without maintaining contact with the body. More details can be found in Appendix~\ref{sec:sam3d-analysis}.

These results answer \textbf{RQ3}: the two deployment pathways expose a tradeoff between deployment speed and contact performance. Direct execution enables rapid, training-free physical inspection of generated motions, while policy learning requires additional data collection and training but achieves substantially higher contact coverage. Together, they suggest a workflow in which generated behaviors can first be inspected through direct execution before investing in policy training when more reliable contact is needed.

\section{Conclusion}
\label{sec:conclusion}
We present \pipeline, an agentic generative simulation framework that converts natural-language task descriptions into pHRI scenarios through coordinated generation, critique, and revision. We generate 50 assistive scenarios from simple prompts without manual intervention. \pipeline{} supports both training-free direct execution and policy learning from simulated demonstrations. In a 12-participant study, learned policies achieved approximately 80\% target completion on bathing and scratching without instructed movement, with no real-world fine-tuning. Participants also rated generated and learned behaviors as consistent with task prompts. These results demonstrate a path from textual pHRI task descriptions to full scenarios and physical robot behavior while reducing task-specific scene and motion authoring.

\paragraph{Limitations and Future Work}
Our real-world evaluation covers two tasks with 12 participants on one robot, leaving broader transfer to other tasks and embodiments for future work. Within the evaluated tasks, maintaining an accurate observation of the human remains important for closed-loop interaction, so multi-view fusion could provide more complete observations when the robot occludes the body. Human movement presents a related challenge, as reflected in the lower completion rate when participants moved their arms during the scratching task. Although our simulator already supports ambient human motion, we do not currently include it during policy training. Collecting demonstrations with moving humans could therefore improve adaptation during real-world interaction.

The generated demonstrations are also limited by the fidelity of simulated contact. Soft-body human simulation exceeded our available computational budget, but could better capture how the body deforms under contact. Combining this simulation capability with force observations, which our current policies do not use, could improve learned contact behavior in the real world.

\bibliographystyle{ACM-Reference-Format}
\bibliography{references}

\clearpage
\appendix

\begin{table*}[t]
\centering
\footnotesize
\setlength{\tabcolsep}{4pt}
\renewcommand{\arraystretch}{1.1}
\caption{Generator input/output, critic input/criteria, and report gates/flags throughout the four generation stages.}
\label{tab:stages}
\begin{tabular}{@{}L{1.5cm}L{3.75cm}L{3.75cm}L{3.75cm}L{4.25cm}@{}}
\toprule
& \textbf{1 Human generation} & \textbf{2 Human placement} & \textbf{3 Robot placement} & \textbf{4 Motion generation} \\
\midrule
\textbf{Generator input} &
\begin{cellist}
\item Two-stage prompt: qualitative pose description, then joint angles via few-shot description--pose pairs
\item Body-shape seed
\end{cellist} &
\begin{cellist}
\item Top-down render of the empty furniture
\item Furniture-frame coordinate rules (normalized $x,y$; head and facing directions)
\end{cellist} &
\begin{cellist}
\item API: \texttt{get\_joint\_pos} over 11 body landmarks
\item Constraints: $x$-axis toward the workspace, camera head-on, yaw-only, base outside the person
\end{cellist} &
\begin{cellist}
\item API: \texttt{get\_joint\_pos} over 12 landmarks; surface projection of a landmark onto the body point cloud; per-point normals; camera pose
\item Waypoint schema and contact constraints
\end{cellist} \\
\addlinespace
\textbf{Generator output} &
\begin{cellist}
\item SMPL-X parameters for body shape
\item SMPL-X angles for hips, knees, neck, shoulders, elbows
\item Posture class (sitting / lying)
\end{cellist} &
\begin{cellist}
\item $(x,y)$ in the furniture frame
\item head direction
\item facing direction
\end{cellist} &
\begin{cellist}
\item Python \texttt{get\_robot\_placement(human)} returning base position and yaw
\end{cellist} &
\begin{cellist}
\item Python \texttt{generate\_trajectory(\ldots)} returning waypoints
\item Each waypoint: position, $3{\times}3$ orientation, velocity, force $\in\{0,1\}$, planner $\in\{$RRT, Cartesian$\}$
\end{cellist} \\
\addlinespace
\textbf{Critic input} &
\begin{cellist}
\item Top-down and axonometric renders
\item Current joint angles
\item History
\end{cellist} &
\begin{cellist}
\item Top-down render (primary); isometric render to confirm or dismiss intersections
\item Current placement parameters
\item History
\end{cellist} &
\begin{cellist}
\item Top-down and isometric renders of scene, human and robot
\item Current code
\item Tagged metrics
\item History
\end{cellist} &
\begin{cellist}
\item Top-down and isometric trajectory renders over the body; rollout video
\item Current code
\item Tagged metrics incl.\ per-waypoint reachability
\item History
\end{cellist} \\
\addlinespace
\textbf{Critic checklist} &
\begin{cellist}
\item Implausible bending or twisting
\item Broken limb configuration
\item Any self-intersection
\item Pose-task mismatch
\end{cellist} &
\begin{cellist}
\item Body-furniture relation matches the task
\item No body-furniture intersection (limbs may hang off when lying; near the edge when seated)
\item Near the edge of wide furniture so a robot can reach
\item Whether $x$, $y$, head or facing should change
\end{cellist} &
\begin{cellist}
\item Proximity: not too far to reach the workspace
\item Orientation: faces the interaction area head-on
\item Task suitability
\end{cellist} &
\begin{cellist}
\item Contact velocity safe
\item Force only on segments that start and end in contact
\item RRT in free space, Cartesian in contact
\item Every waypoint reachable
\item Contact waypoints cover the body region the task names
\end{cellist} \\
\addlinespace
\textbf{Report: gates} &
\begin{cellist}
\item None
\end{cellist} &
\begin{cellist}
\item None
\end{cellist} &
\begin{cellist}
\item G: base tilt $\leq 1^\circ$ after settling
\end{cellist} &
\begin{cellist}
\item G: schema assertions (force, planner, contact velocity $<0.05$\,m/s, first waypoint force-free)
\item G: all waypoints reachable
\end{cellist} \\
\addlinespace
\textbf{Report: advisory flags} &
\begin{cellist}
\item F: self-collision pairs $>0$
\item F: max penetration $\geq 1$\,cm
\end{cellist} &
\begin{cellist}
\item None
\end{cellist} &
\begin{cellist}
\item None
\end{cellist} &
\begin{cellist}
\item F: contact waypoints off target
\item F: surface-normal angle $>75^\circ$
\item F: contact segment $>5$\,cm from its landmark
\item F: goal-observation distance mean $>5$\,cm / max $>10$\,cm
\end{cellist} \\
\bottomrule
\end{tabular}
\end{table*}

\section{Stage-Specific Critic Criteria and Metrics}
\label{sec:stage-breakdown}

Table~\ref{tab:stages} summarizes the four scenario-generation
stages. Orchestration also extends through demonstration collection,
data processing, policy training, and simulated evaluation, using
the same structured stage reports. Table~\ref{tab:policy-stage-reports}
summarizes their gates and advisory flags.

Training reports assess completion, numerical stability, and
validation performance. Policy rollouts assess task competence
on training and held-out human bodies. These findings guide the
orchestrator's recovery decisions. For example, worsening validation
loss can motivate selecting an earlier checkpoint, while an
episode-level outlier can motivate revisiting its demonstrations.
Evaluation failures can similarly prompt changes to policy training
or upstream scenario generation. Revising demonstrations invalidates
dependent training and evaluation results.

\begin{table*}[t]
\centering
\footnotesize
\setlength{\tabcolsep}{4pt}
\renewcommand{\arraystretch}{1.15}
\caption{Stage reports for the policy-learning branch.
Gates specify required conditions; advisory flags identify concerns
for orchestrator review. The same training criteria apply separately
to the high- and low-level policies.}
\label{tab:policy-stage-reports}
\begin{tabular}{@{}L{2.2cm}L{5.5cm}L{8.3cm}@{}}
\toprule
\textbf{Stage} & \textbf{Gates} & \textbf{Advisory flags} \\
\midrule

Data collection &
No nonfinite values in training data. &
\begin{cellist}
    \item Dropped-episode fraction $\geq 0.2$
    \item Episode-length ratio outside $[0.8,1.2]$
    \item Mean tracking error $>5$\,cm or maximum $>15$\,cm
\end{cellist} \\

\addlinespace
Policy training 
&
\begin{cellist}
    \item Planned epochs completed
    \item No training or validation batches with nonfinite loss
\end{cellist}
&
\begin{cellist}
    \item Final validation loss $>1.5\times$ the best epoch's loss
    \item Reduction from initial to best validation loss $<50\%$
    \item Worst held-out episode loss $>3\times$ the median episode loss
\end{cellist} \\

\addlinespace
Simulated evaluation &
\begin{cellist}
    \item Every configured human produces an evaluation episode
    \item Success rates on training and held-out bodies each meet
the configured minimum
\end{cellist}
&
\begin{cellist}
    \item Training-body success rate minus held-out success rate $>0.25$
    \item Mean goal error $>5$\,cm
    \item Stalled-episode fraction $>0.2$
    \item Worst subgoal completion $<0.5$
    \item Success-rate spread across humans $>0.5$
    \item Visual-numerical success agreement $<0.7$
\end{cellist} \\

\bottomrule
\end{tabular}
\end{table*}

\section{Generated Scenarios in Simulation}
\label{sec:sim-details}

The benchmark suite consists of 50 physical human-robot interaction (pHRI) scenarios spanning nine caregiving activity families, designed to systematically evaluate robot trajectory planning across realistic care settings. Across these families, scenarios vary in target body segment, body posture (lying versus sitting), spatial environment (drawn from 34 distinct room layouts), and natural-language phrasing. Over half of the scenarios target limbs and narrow body segments rather than broad planar surfaces, testing trajectory generation on small, highly curved geometry under realistic spatial constraints. Table~\ref{tab:family_summary} summarizes the distribution and target scope across all nine activity families.

Generating the complete 50-scenario suite required a simulation time of $254.4\text{ GPU-hours}$ and $9{,}170$ LLM calls across $1{,}209$ total stage attempts without human intervention. Across all pipeline stages, generation averaged approximately $183.4\text{k}$ prompt input tokens and $12.1\text{k}$ output tokens per completed scenario, reaching an aggregate suite footprint of $9.17\text{M}$ input and $605\text{k}$ output tokens. Utilizing aggregate token footprints rather than monetary API costs provides a stable, price-invariant benchmark of computational complexity. 

\begin{table}[h]
  \centering
  \caption{Summary of the 50 generated caregiving scenarios grouped by activity family, detailing scenario counts and primary target body regions.}
  \label{tab:family_summary}
  \small
  \begin{tabular}{@{}lcl@{}}
    \toprule
    Activity Family & Count & Primary Target Body Regions \\
    \midrule
    Back percussion    & 7 & Upper back, mid back, lower back \\
    Itch relief        & 7 & Mid/lower back, scalp, shoulder, arm, leg \\
    Sponge bathing     & 6 & Head/neck, foot, hand, back, leg, arm \\
    Lotion application & 5 & Arm/hand, back, leg, shoulder/neck, hips \\
    Skin assessment    & 5 & Heel, hips, spine, shoulder, elbow \\
    Hand warming       & 5 & Palm, back of hand, knuckles, wrist \\
    Towel drying       & 5 & Arm, back, foot, head/neck, hip/thigh \\
    Comforting touch   & 5 & Shoulder, hand, back, arm, head \\
    Edema massage      & 5 & Leg, arm/hand, shoulder, calf, foot \\
    \bottomrule
  \end{tabular}
\end{table}

\section{Direct Execution Analysis}
\label{sec:sam3d-analysis}

\begin{figure*}[tb]
  \centering
  \includegraphics[width=0.8\linewidth]{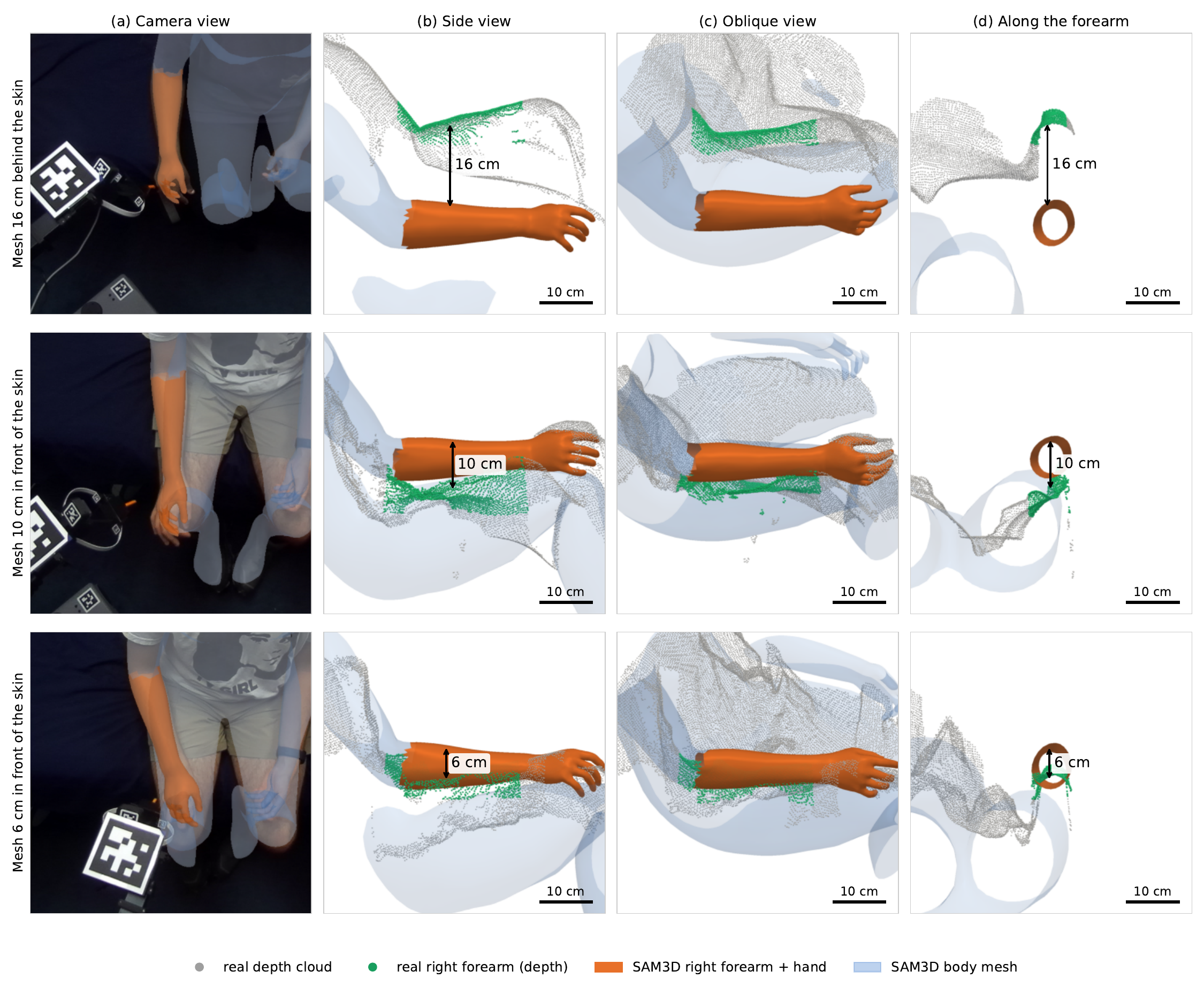}
  \caption{Illustrations of Fast SAM3D Body fitting results, compared against the actual human point cloud.}
  \label{fig:sam3d}
\end{figure*}

Direct execution transfers generated trajectories into the real world by fitting a single-frame SMPL-X human body mesh using Fast SAM 3D Body, enabling open-loop execution without demonstration collection or policy training. However, single-view mesh recovery inherently suffers from depth ambiguity along the camera's viewing ray, as depth information is not taken into account in this model. This issue is particularly relevant for the scratching task. As illustrated in Figure~\ref{fig:sam3d}, while the 2D mesh projection aligns cleanly with the human's silhouette from the primary camera perspective (Column a), multi-angle visualizations (Columns b--d) reveal substantial depth discrepancies along the camera viewing direction. Across representative testing frames, the fitted forearm lies $16\text{ cm}$ behind the actual arm surface in Row 1, and $10\text{ cm}$ and $6\text{ cm}$ in front of the arm surface in Rows 2 and 3, respectively.

These depth offsets motivated restricting direct execution to the bathing task and excluding it from the forearm scratching evaluation. First, because a human forearm is narrow ($5\text{--}8\text{ cm}$ wide), depth errors of $6\text{--}16\text{ cm}$ displace planned contact waypoints entirely off the limb into open space or deep into the arm. While bathing tolerates positional offsets due to broad target surfaces (back and legs) and towel compliance, localized scratching fails completely under open-loop control. Second, when the fitted mesh lies behind the physical skin (e.g., Row 1), open-loop trajectory execution commands the end-effector to press directly into the human's arm. Since there is no closed-loop contact feedback to interrupt motion, deploying a scratching tool under these conditions introduces a physical safety hazard.

\section{Hardware Setup}
\label{sec:hardware}
We deploy our policies on a Hello Robot Stretch 3 mobile manipulator equipped with a task-specific end-effector (e.g., a forearm scratcher or bathing towel) and a ceiling-mounted ZED 2i camera. To calibrate the transform between the robot and the camera, we place an ArUco tag on the robot's wrist joint. A separate GPU workstation runs the learned high- and low-level policies for inference, communicating wirelessly with the robot's onboard computer to stream observations and commands. The workstation segments the human from the RGB image using SAM 3~\cite{carion2025sam}, constructs a point cloud from the masked depth image, and queries the hierarchical policy to produce end-effector goals and actions that are converted to joint commands and sent to the robot for execution.

\section{Policy Training}
\label{sec:policy-details}

\paragraph{Demonstration collection.}
For each task, we collect 1,000 demonstrations across 100 generated human bodies, varying body pose and shape. Across executions, we randomize the following parameters for generalization:
\begin{itemize}[
  topsep=0pt,      
  partopsep=0pt,   
  parsep=0pt,      
  itemsep=0.1em,   
  leftmargin=1.5em,  
  labelsep=0.5em   
]
  \item Human body shape: $\beta$ parameters of the SMPL-X model are randomly sampled from the standard Gaussian distribution---$\beta\sim\mathcal{N}(0,I)$.
  \item Camera pose: Nominally the camera is placed directly above the human center at a height of 2.5\,m, looking straight down at the human center. We apply perturbations to the camera position and the camera ``lookat'' position as follows:\begin{itemize}[
      topsep=0pt,      
      partopsep=0pt,   
      parsep=0pt,      
      itemsep=0.1em,   
      leftmargin=1.5em,  
      labelsep=0.5em   
    ]
    \item $\Delta x_{\text{cam}} \sim \text{Unif}([-0.1, -0.1, -0.7], [0.1, 0.1, 0])$
    \item $\Delta x_{\text{lookat}} \sim \text{Unif}([-0.1, -0.1, -0.1], [0.1, 0.1, 0.1])$
  \end{itemize}
  \item Initial robot base pose: Perturbations are applied to the generated robot base position and base yaw as follows:\begin{itemize}[
      topsep=0pt,      
      partopsep=0pt,   
      parsep=0pt,      
      itemsep=0.1em,   
      leftmargin=1.5em,  
      labelsep=0.5em   
    ]
    \item $\Delta x_{\text{robot}} \sim \text{Unif}([-0.04, -0.04, 0], [0.04, 0.04, 0.15])$
    \item $\Delta\theta_{\text{robot}} \sim \text{Unif}(-4^\circ, 4^\circ)$
  \end{itemize}
  \item Initial robot joint configuration: These are picked specifically for a Stretch 3 robot:\begin{itemize}[
      topsep=0pt,      
      partopsep=0pt,   
      parsep=0pt,      
      itemsep=0.1em,   
      leftmargin=1.5em,  
      labelsep=0.5em   
    ]
      \item $q_{\text{lift}} \sim \text{Unif}(0.75\,\text{m}, 0.95\,\text{m})$
        \item $q_{\text{arm},i} \sim \text{Unif}(0\,\text{m}, 0.01\,\text{m}), \quad i = 1, \dots, 4$
        \item $q_{\text{wrist yaw}} \sim \text{Unif}(-20^\circ, 20^\circ)$
        \item $q_{\text{wrist pitch}} \sim \text{Unif}(-27.2^\circ, -7.2^\circ)$
        \item $q_{\text{wrist roll}} \sim \text{Unif}(-30^\circ, 30^\circ)$
  \end{itemize}
\end{itemize}
Each observation includes a segmented human point cloud downsampled to 4,500 points using farthest point sampling, the current end-effector pose, and the corresponding waypoint and action labels. During preprocessing, spatial observations and goal poses are expressed relative to the human point-cloud centroid.

\paragraph{Trajectory augmentation.}
We retain each original demonstration and generate 16 additional perturbed copies. At each timestep, we perturb the end-effector pose and recompute an eight-step action sequence that moves from the perturbed pose toward the demonstrated trajectory. Translation perturbations are sampled from a zero-mean Gaussian with a standard deviation of 2\,cm per coordinate and clipped to $\pm5$\,cm. Rotation perturbations use a random axis and a Gaussian angle with a standard deviation of $0.175$\,rad, clipped to $\pm0.524$\,rad. During contact, tangential translations are scaled by $0.5$, translations along the local surface normal are restricted to the outward direction, and rotation perturbations are scaled by $0.3$. Goal poses and waypoint indices are reassigned using the nearest demonstrated end-effector position within five timesteps of the source frame. This preserves consistency between perturbed observations, motion stages, and corrective-action targets.

\paragraph{Observation augmentation.}
We additionally apply online noise during training. For the high-level policy, independent Gaussian noise is added to each point-cloud coordinate with a standard deviation of 1\,mm and clipped to $\pm3$\,mm. Each training frame is also presented with two additional end-effector pose perturbations, using translation and rotation-vector standard deviations of 4\,mm and $0.15$\,rad per coordinate, respectively, while retaining its goal label. For the low-level policy, point-cloud coordinates receive Gaussian noise with a standard deviation of 3\,mm. Together, these augmentations expose the policies to observation noise and deviations from the demonstrated end-effector trajectory.

\paragraph{Training setup.}
We train separate hierarchical policies for bathing and scratching using PyTorch distributed data-parallel training on four NVIDIA RTX PRO 5000 Blackwell GPUs, each with 48\,GB of memory. The high-level Point Transformer V3 policy is trained for 15 epochs with a batch size of 128 per GPU. We use AdamW with weight decay $5\times10^{-3}$ and a cosine OneCycle learning-rate schedule, with peak learning rates of $10^{-3}$ for non-backbone parameters and $10^{-4}$ for backbone blocks. The low-level diffusion policy is trained with demonstration goal poses as conditioning inputs and an eight-step action prediction horizon. It uses AdamW with a learning rate of $10^{-4}$, weight decay $10^{-6}$, a cosine schedule with 500 warm-up steps, and a batch size of 256 per GPU. We train the low-level policy for 30 epochs for bathing and 20 epochs for scratching. All training uses simulated demonstrations, with no real-world fine-tuning.

\paragraph{Training time.}
Table~\ref{tab:policy-training-time} reports approximate wall-clock durations for the completed training runs. These measurements include training and validation but exclude scenario generation, demonstration collection, preprocessing, and earlier training attempts. Training both policy levels takes approximately 20 hours per task on the four-GPU workstation.

\begin{table}[H]
    \centering
    \caption{Approximate policy training time on four NVIDIA RTX PRO 5000 Blackwell GPUs.}
    \label{tab:policy-training-time}
    \begin{tabular}{lrrr}
        \hline
        Task & High-level (h) & Low-level (h) & Total (h) \\
        \hline
        Bathing    & 17.5 & 2.3 & 19.8 \\
        Scratching & 17.7 & 1.2 & 19.0 \\
        \hline
    \end{tabular}
\end{table}

\section{User Study Task Prompts}
\label{sec:task-prompts}
Bathing: "A person lies face-down on a bed, arms resting alongside the torso and legs extended straight. A robot bathes them with a sponge, wiping the whole back in several passes and then the back of each leg down to the ankle, and it pulls the sponge away from the body when it finishes. The robot must not touch the buttocks or the area between the legs."

Scratching: "A person sits in a chair with their right forearm held slightly forward. A robot stands in front of the person and slightly to their right. Keeping its gripper arm separated from the person, it reaches down with the scratcher and gently scratches the top of the full right forearm, back and forth three times, maintaining light contact. Once done, it retreats."

\end{document}